\documentclass[letterpaper]{article} 

\usepackage{aaai2026}  
\usepackage{times}  
\usepackage{helvet}  
\usepackage{courier}  
\usepackage[hyphens]{url}  
\usepackage{graphicx} 
\usepackage{natbib}  
\usepackage{caption} 
\usepackage{algorithm}
\usepackage{algorithmic}
\usepackage[utf8]{inputenc} 
\usepackage{newfloat}
\usepackage{multirow} 
\usepackage{listings}
\DeclareCaptionStyle{ruled}{labelfont=normalfont,labelsep=colon,strut=off} 
\floatstyle{ruled}
\newfloat{listing}{tb}{lst}{}
\floatname{listing}{Listing}
\usepackage{amssymb}

\usepackage{booktabs}
\usepackage{multirow}
\usepackage[most]{tcolorbox} 
\usepackage{float}  
\usepackage{adjustbox}
\usepackage{caption}
\usepackage[table,xcdraw]{xcolor}
\usepackage{multirow}
\usepackage{enumitem}
\title{AI-Powered Early Diagnosis of Mental Health Disorders from Real-World Clinical Conversations}

\title{AI-Powered Early Diagnosis of Mental Health Disorders from Real-World Clinical Conversations}
\author{
  Jianfeng Zhu\textsuperscript{\rm 1},
  Julina Maharjan\textsuperscript{\rm 1},
  Xinyu Li\textsuperscript{\rm 1},
  Karin G. Coifman\textsuperscript{\rm 2},
  Ruoming Jin\textsuperscript{\rm 1}
}
\affiliations{
 \textsuperscript{\rm 1} Department of Computer Science, Kent State University \\
\textsuperscript{\rm 2} Department of Psychological Sciences, Kent State University \\
\{jzhu6, jmaharja, xli5, rjin\}@kent.edu, kcoifman@kent.edu
}

\nocopyright
\begin{document}

\maketitle

\begin{abstract}
Mental health disorders remain among the leading cause of disability worldwide, yet conditions such as depression, anxiety, and Post-Traumatic Stress Disorder (PTSD) are frequently underdiagnosed or misdiagnosed due to subjective assessments, limited clinical resources, and stigma and low awareness. In primary care settings, studies show that providers misidentify depression or anxiety in over 60\% of cases, highlighting the urgent need for scalable, accessible, and context-aware diagnostic tools that can support early detection and intervention.
In this study, we evaluate the effectiveness of machine learning models for mental health screening using a unique dataset of 553 real-world, semi-structured interviews, each paried with ground-truth diagnoses for major depressive episodes (MDE), anxiety disorders, and PTSD. We benchmark multiple model classes, including zero-shot prompting with GPT-4.1 Mini and Meta-LLaMA, as well as fine-tuned RoBERTa models using Low-Rank Adaptation (LoRA). Our models achieve over 80\% accuracy across diagnostic categories, with especially strong performance on \textit{PTSD} (up to 89\% accuracy and 98\% recall).
We also find that using shorter context, focused context segments improves recall, suggesting that focused narrative cues enhance detection sensitivity. LoRA fine-tuning proves both efficient and effective, with lower-rank configurations (e.g., rank 8 and 16) maintaining competitive performance across evaluation metrics. Our results demonstrate that LLM-based models can offer substantial improvements over traditional self-report screening tools, providing a path toward low-barrier, AI-powerd early diagnosis. This work lays the groundwork for integrating machine learning into real-world clinical workflows, particularly in low-resource or high-stigma environments where access to timely mental health care is most limited.

\end{abstract}

\begin{links}
    \link{Code}{https://anonymous.4open.science/r/AAAI2026_Depression1-E152/}
\end{links}
\section{1. Introduction}
Mental health disorders account for four of the ten leading cause of disability worldwide \cite{alhamed2024using,ben2025assessing,world2001world}. Individuals affected by these conditions experience disproportionally high rates of adverse health behaviors, including tobacco smoking, substance use, physical inactivity, and poor diet, which contribute to elevated risks of chronic medical illness and premature mortality \cite{goodell2011mental,laursen2014excess}. Despite their global burden, mental health disorder remain underdiagnosed, undertreated and stigmatized, particularly in resource constrained or high barrier settings \cite{world2001world}.

A major contributor to this treatment gap lies in the limitations of traditional diagnostic tools. Mental health assessments rely on subjective measurements, such as structured interviews and self-reports questionnaires \cite{tennyson2016challenges}. While these instruments are cost-effective and scalable, they suffer from social desirability bias, inaccurate recall, and limited interoperability of symptoms, particularly in cognitively impaired or stigmatized populations \cite{haberer2013furthering}. Moreover, timely and accurate diagnosis is complicated by the high comorbidity of condition such as depression, anxiety and post-traumatic stress disorder (PTSD), which often present overlapping symptoms and are difficult to disentangle in practice. This diagnostic ambiguity, compounded by limited clinical resources, contributes to widespread misdiagnosis and under-recognition in both high- and low-income settings \cite{auxemery2018post}. 

These challenges underscore the urgent need for scalable, proactive, and context aware screening tools that can identify at-risk individuals earlier and more accurately \cite{tennyson2016challenges}. In response to these demands, 
traditional machine learning methods such as SVM, random forest, etc ~\cite{saidi2020hybrid, cacheda2019early, islam2018depression} have long been extensively studied and applied to detect various mental health conditions by leveraging structured inputs and hand-crafted features derived from linguistic, behavioral, or physiological signals. While these approaches have shown promise, they often result in suboptimal performance due to the limited expressiveness~\cite{van2023not}. Subsequent advances in deep learning enabled end-to-end models that learn representations directly from raw text or speech data, offering improved performance and flexibility \cite{su2020deep}. However, these models still typically require large labeled datasets, suffer from poor interpretability, and remain sensitive to domain shifts ~\cite{ramasamy2018recent}. 
Recent advances in Large Language Models (LLMs) offer a compelling new direction. By analyzing naturalistic user input, such as interview transcripts and spontaneous conversations, LLMs can potentially infer early signs of depression, anxiety, and other mental health disorders~\cite{xu2024mental}. These models require no clinical infrastructure and can be deployed in low touch, user facing environments, enabling broad access and repeated monitoring. This study takes a step in that direction by asking a core question: \textit{Can we harness the broad prior knowledge and strong text comprehension skills of large language models to detect early signs of mental disorders from naturalistic conversations, without relying on expert-crafted features or clinical annotations?} 
\begin{figure}[H]
    \centering
    \includegraphics[width=0.4\textheight, keepaspectratio]{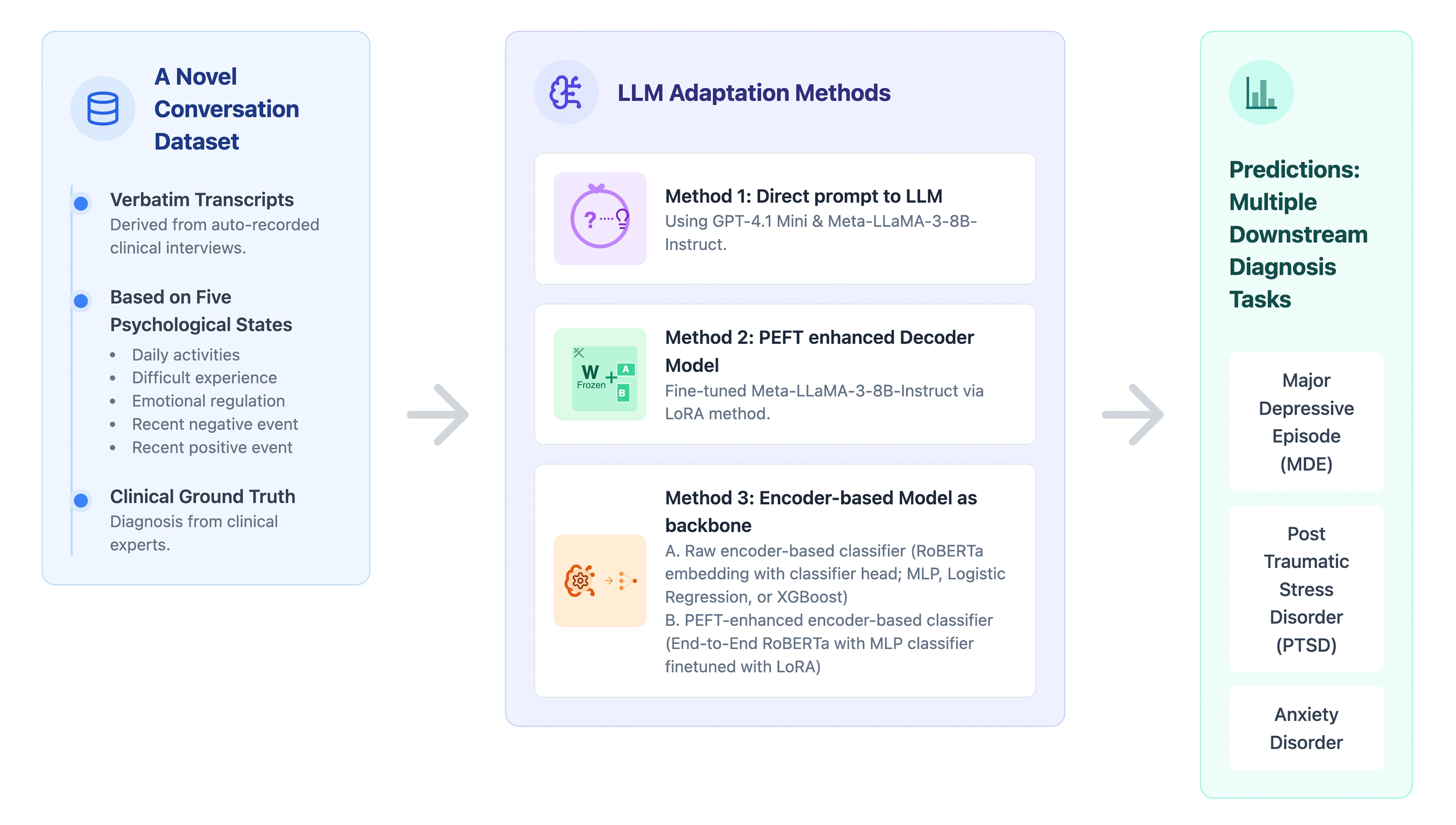}
    \caption{Overview of dataset, LLM adaptation methods, and prediction targets.}
    \label{fig:overview_pipeline}
\end{figure}

Unlike prior work which based on self-reported surveys or social media data on single mental health disoders \cite{bucur2024leveraging,zhu2024user,sarabadani2025pkg,bartal2024chatgpt}. We leverage a unique dataset of real-world, semi-structured psychiatric interviews with ground-truth clinical diagnoses assessed within the same time window. This setting more closely reflects the nuance and variability of real clinical conversations. We formalize the task as multi-label text classification, where given an interview transcript, the model must predict the presence of \textbf{(1) major depressive episodes (MDE)}, \textbf{(2) PTSD}, and \textbf{(3) anxiety disorders}.

We evaluate both encoder-based and decoder-based language models, with optional enhancement via parameter-efficient fine-tuning (PEFT) adapters such as LoRA. Decoder-based models (e.g., GPT-4.1, Meta-LLaMA) are assessed in zero-shot settings, while encoder-based models are optionally fine-tuned using embedding-based classifiers with or without LoRA augmentation. This design allows for a rigorous, side-by-side comparison of general-purpose versus customized diagnostic strategies under realistic data and deployment constraints.An overview of our modeling pipeline is illustrated in Figure ~\ref{fig:overview_pipeline}.

\section{2. Related Work}
Prior work in mental health assessment has largely relied on standardized self-report instruments designed to evaluate specific psychiatric conditions. For instance, the Patient Health Questionnaire (PHQ-9) is a widely used self-report scale for assessing major depressive disorder, based on DSM-IV criteria \cite{kroenke2001patient}. Similarly, the GAD-7 is a brief, clinically validated measure for evaluating the severity of generalized anxiety disorder \cite{spitzer2006brief}, and the PTSD Checklist for DSM-5 (PCL-5) captures core symptoms of post-traumatic stress disorder across four diagnostic clusters, in alignment with DSM-5 standards \cite{blevins2015posttraumatic}.\\
While these instruments are cost-effective and easy to administer, they suffer from several important limitations. First, each instrument typically targets only one disorder per administration, despite increasing evidence that conditions such as depression, anxiety, and PTSD frequently co-occur in clinical populations \cite{lai2019prevalence,hawkins2009tale,devylder2014co}. Second, because these tools rely entirely on self-reported data, they are vulnerable to social desirability bias, recall inaccuracies, and limited self-awareness, especially among vulnerable or cognitively impaired individuals.  Finally, in resource-limited settings or during health emergencies (e.g., pandemics), access to trained professionals or structured screening processes may be severely constrained, leaving many individuals undiagnosed or misdiagnosed \cite{kumar2025evaluation,alhalaseh2021allocation}. For example, a synthesis of 157 studies found that only one in three individuals with mild depression is correctly identified in primary care \cite{mitchell2011can}, and a meta-analysis of 41 high-quality studies, encompassing over 50,000 patients, found that general practitioners correctly identified depression in just 47.3\% of cases. Notably, these studies also found that false positives often outnumbered true positives, and a substantial proportion of cases were entirely missed \cite{mitchell2009clinical}.\\
In response to the limitations of self-report instruments, recent research has explored machine learning (ML) and natural language processing (NLP) approaches to automatically detect mental health conditions from diverse data sources, including text inputs, questionnaires, and social data \cite{wshah2019predicting,le2021machine,chiong2021textual,maharjan2025differential}. For instance, Priya et al. study applied ML algorithms to classify individuals across five severity levels of anxiety, depression, and stress, based on questionnaire responses. While Naïve Bayes achieved the highest accuracy, Random Forest was identified as the overall best-performing model \cite{priya2020predicting,nemesure2021predictive}. Other studies have demonstrated the feasibility of using ML methods to identify PTSD, leveraging structured assessments, language-based emotion regulation features, and treatment selection models \cite{christ2021machine,held2022will,sawalha2022detecting,9120899}. Despite these promising directions, the implementation of digital mental health solutions still faces substantial challenges—particularly in terms of evaluation rigor and practical effectiveness \cite{balcombe2021digital}.\\ 
Unlike traditional machine learning methods that often require large amounts of task-specific training data and manually engineered features, large language models (LLMs) operate under a fundamentally different paradigm. Powered by transformer architectures and the self-attention mechanism \cite{vaswani2017attention}, LLMs are pretrained on massive text corpora and can perform zero-shot or few-shot inference based solely on textual prompts\cite{wang2023largelanguagemodelszeroshot,hasan2024zerofewshotpromptingllms}. This shift enables LLMs to understand rich contextual cues directly from free-form natural language—without the need for structured input formats, disorder-specific questionnaires, or annotated training data \cite{zhu2024user,trap7969,zhu2025leveraginglargelanguagemodels}. \\
Several recent studies have demonstrated the promise of LLMs in clinical mental health applications. For example, GPT-4 was shown to infer social anxiety symptom severity from semi-structured clinical interviews, achieving a correlation of r = 0.79 with validated self-report measures \cite{ohse2024gpt4}. In the context of depression diagnosis, MDD-LLM (70B) achieved an accuracy of 0.8378 and AUROC of 0.8919, significantly outperforming conventional machine and deep learning approaches \cite{SHA2025119774}. Another study on PTSD detection used few-shot prompting with DSM-5 criteria to achieve AUROC = 0.737, with performance varying by symptom severity and comorbid depression status \cite{chen2025detectingptsdclinicalinterviews}. These findings underscore the capacity of LLMs to deliver clinically meaningful insights from minimally structured data making them especially valuable for scalable mental health screening in low-resource or underserved contexts.\\
Building on this progress, our work makes several key contributions. We evaluate both general-purpose LLMs (e.g., GPT-4, Meta-LLaMA) and parameter-efficient fine-tuned models (e.g., RoBERTa with LoRA) on a unique dataset of 555 real-world psychiatric interviews. 


\section{3. Datasets}
We utilize a novel dataset comprising 555 U.S. adults, collected across multiple behavioral research studies investigating individual responses to transitional or adverse life events, such as occupational stress, chronic illness, or traumatic exposure. All participants provided written informed consent under IRB-approved protocols. Table~\ref{tab:demographics} summarizes the participant demographics. The dataset includes 553 individuals, with a balanced gender distribution (278 male, 275 female), and a majority identifying as White (431). Age is broadly distributed across five brackets, and most participants report some college or higher education. This diversity supports robust downstream analysis. During these interviews, all participants answered the same five questions in a predetermined sequence. The first question required participants to describe their activities from waking to sleep on the previous day. The subsequent questions explored their personal experiences with a recent challenging event or situation, their coping strategies for that challenge, an unrelated recent unpleasant event, and, finally, a recent positive experience.\\

\begin{table}[htbp]
\centering
\begin{tabular}{llr}
\toprule
\textbf{Category} & \textbf{Subgroup} & \textbf{Count} \\
\midrule
\multirow{5}{*}{Age} 
  & $<$25         & 123 \\
  & 25--34        & 128 \\
  & 35--44        & 96  \\
  & 45--59        & 132 \\
  & 60+           & 74  \\
\midrule
\multirow{2}{*}{Sex} 
  & Male          & 278 \\
  & Female        & 275 \\
\midrule
\multirow{2}{*}{Race} 
  & White         & 431 \\
  & Other         & 122 \\
\midrule
\multirow{4}{*}{Education} 
  & High school or below & 98 \\
  & Some college          & 293 \\
  & College or above      & 134 \\
  & Unknown               & 28 \\
\bottomrule
\end{tabular}

\caption{Participant Demographics Summary}
\label{tab:demographics}
\end{table}

Interviewers were trained to provide standardized prompts, and participants were encouraged to speak spontaneously for up to three minutes in response to each question. Each interview session resulted in approximately 15 minutes of recorded speech per participant. The speech was then recorded and transcribed according to convention and recommendations \cite{coifman2007repressive,coifman2010distress,coifman2016context,harvey2014emotion}.

Each participant's text responses were paired with their corresponding depressive symptoms, derived from Structured Clinical Interview for DSM (SCID) reports . Demographic information, including age and sex, was collected across studies and harmonized into a unified format. The mean participant age was 39.36 years (SD = 16.0). The sample was approximately gender-balanced: 278 participants identified as female, 277 as male, and one participant did not report their sex.\\
The average length of the full interviews was approximately 2,955 words ($SD = 1{,}855$). This setup allows for a fine-grained analysis of how narrative depth influences psychiatric signal extraction across model types. For foundation models constrained by input length limitations (e.g., Meta-LLaMA-3-8B), we adopt a chunk-based inference strategy, whereby each user transcript is segmented into overlapping chunks of 512, 1024, or 2048 tokens. Model predictions are computed for each chunk independently and averaged to obtain a user-level binary decision. This design enables scalable prediction while preserving diagnostic signal across extended narrative contexts.\\

To illustrate the richness of the narrative content, Table~\ref{tab:interview_examples} provides representative excerpts from a single participant across the five prompt domains. These responses reflect the emotional nuance and thematic complexity of the dataset, presenting a realistic and ecologically valid benchmark for naturalistic mental health inference.

\begin{table}[t]
\centering
\resizebox{0.48\textwidth}{!}{%
\begin{tabular}{p{3.3cm} p{7.5cm}}
\toprule
\textbf{Prompt} & \textbf{Response Excerpt} \\
\midrule
Daily Activities & ``Beginning of the day, uh I have two sons, one is [xxxx]... played outside... gym...'' \\
Difficult Experience & ``Being a firefighter... challenging and amazing experience... bad experiences too...'' \\
Emotion Regulation & ``You talk to people you trust at work... My wife and I have been married...'' \\
Negative Event & ``First big incident... stressful leading into it...'' \\
Positive Event & ``First baby we delivered—early morning call... heroin case...'' \\
\bottomrule
\end{tabular}%
}
\caption{Example Interview Responses (Participant ID: 001)}
\label{tab:interview_examples}
\end{table}

\section{4. Methodology}
In this section, we systematically explain the variaous methods to evaluate the abilities of modern AI models in mental health disorder domain.
\subsection{Method 1: Multi-Disorder Inference via Direct Prompting}
First, we investigate whether modern foundation models can infer multiple psychiatric disorders directly from raw interview text in a zero-shot setting. Specifically, we prompt large language models (LLMs) to identify signs of psychiatric conditions, including \textbf{depression}, \textbf{PTSD}, and \textbf{anxiety} without any task-specific training or fine-tuning. We examine two state-of-the-art LLM models:
\begin{itemize}
    \item \textbf{GPT-4.1 Mini}: Deployed in a zero-shot configuration using custom-designed prompts tailed to each disorder. This setup serves as a baseline for scalable mental health screening without the need for retraining.
    \item \textbf{Meta-LLaMA-3-8B-Instruct}: Evaluated under zero-shot prompting. To alleviate the complexity introduced by long-range dependencies, interview transcripts were further segmented into small chunks of size 512, 1024, or 2048 with a fixed overlapping rate, each chunked transcript was then reformulated into a prompt and subsequently passed to the model for binary classification. Final user-level predictions were derived by averaging prediction scores across all chunks.
\end{itemize}
We employ the following standardized prompt for all models and disorders, adhering to a binary prediction format to ensure clinical interpretability:
\begin{tcolorbox}[width=\linewidth,colback=gray!5!white, colframe=gray!50!black, title=Prompt: Psychiatric Disorder Inference (LLM-Based)]
\textbf{You are an AI assisting mental health professionals in identifying psychiatric disorders.}\\
\textbf{Input Data:} \\
\texttt{"\{response\_text\}"} — a free-form participant response from a semi-structured interview.\\
\textbf{Task:} \\
Analyze the participant's response and determine whether there are signs of a psychiatric disorder. Do not include any reasoning or explanation.\\
\textbf{Output Format:} \\
Respond with a single-line binary decision:
\begin{quote}
\texttt{Prediction: Yes} \quad \\ \quad \texttt{Prediction: No}
\end{quote}
\end{tcolorbox}

\subsection{Method 2: Decoder-based Model as Backbone with PEFT Enhancement}
To facilitate domain adaptation to mental health tasks, we adopt a parameter-efficient fine-tuning (PEFT) strategy. In particular, we apply \textbf{Low-Rank Adaptation (LoRA)} to align a pretrained transformer-based language model with disorder-specific semantics in a computationally efficient manner. We fine-tune \texttt{Meta-LLaMA-3-8B-Instruct} separately for each psychiatric condition (depression, PTSD, and anxiety) using binary supervision. This enables efficient adaptation with minimal additional parameters while preserving general linguistic knowledge.


\subsection{Method 3: Encoder-based Model as Backbone}
We also evaluate the diagnostic ability of encoder-based language model. Typically, we choose the widely used generic encoder model \texttt{RoBERTa-base}~\cite{liu2019roberta}  and \texttt{
all-roberta-large-v1} through the Sentence-Transformers library ~\cite{reimers-2019-sentence-bert} as our backbone. 
\subsubsection{Adaptation to Long-Form Input}
Unlike decoder-based language models, which typically support large context windows, encoder-based models such as the BERT family are constrained by relatively short input lengths. To enable processing of long-form text, we adopt a 2-step \underline{chunking and aggregation} strategy that first segments input sequences into manageable chunks, and then aggregates their representations to construct a comprehensive embedding for downstream classification. 

\textbf{Step1: Chunking}: Given an user's transcript $x$, consisting of a long sequence of tokens, we split it into overlapping chunks $x_i$ of size $c$ with a fixed overlap ratio. Each chunk is then independently encoded via a PEFT equipped RoBERTa encoder:
\begin{equation}
    h_i = \text{RoBERTa}(x_i)
\end{equation}
We extract the [CLS] token from each chunk embedding, resulting in a chunk-level representation matrix $H = [h_1^{\text{[CLS]}}, \dots, h_T^{\text{[CLS]}}] \in \mathbb{R}^{T \times d}$, where $T$ is the number of chunks and $d$ is the hidden size.

\textbf{Step 2: Aggregation}: Once the chunk-level transcript representation is obtained, we aggregate the high dimensional matrix $H$ into a single vector $h$ as the final user representation. In particular, we opted for two different aggregation strategies:
\begin{itemize}
    \item Mean pooling: $h = \frac{1}{T} \sum_{i=1}^{T} h_i^{\text{[CLS]}}$
    \item Max pooling: $h = \max_{i=1}^{T} h_i^{\text{[CLS]}}$
\end{itemize}

\subsubsection{Raw encoder-based embedding}
As a complementary baseline, we leverage the embeddings by feeding user transcripts to raw encoder model, including RoBERTa-base and all-roberta-large-v1. These embedding representations are then passed to lightweight classifiers—logistic regression, multilayer perceptrons (MLP), and XGBoost—to predict binary disorder labels.

\subsubsection{PEFT-enhanced embedding}
We also leveraged PEFT-method, such as LoRA to fine tune the pretrained encoder, attempting to align the encoder language model with the mental health domain.

To enable end-to-end prediction, we feed the learned text embeddings into a simple classification module designed for depression detection. This module—such as a lightweight MLP—maps the semantic representations to task-specific labels. We also apply layer normalization to stabilize training and include L2 regularization on classifier weights to reduce overfitting..

\section{5. Experiments and Results}
\begin{table*}[t]
\centering 
\small
\setlength{\tabcolsep}{5pt}
\begin{tabular}{llccc|ccc|ccc} 
\toprule
\multirow{2}{*}{Category} & \multirow{2}{*}{Model} & \multicolumn{3}{c|}{\textbf{Depression}} & \multicolumn{3}{c|}{\textbf{PTSD}} & \multicolumn{3}{c}{\textbf{Anxiety}} \\
\cmidrule(lr){3-5} \cmidrule(lr){6-8} \cmidrule(lr){9-11}
& & Acc & Rec & F1 & Acc & Rec & F1 & Acc & Rec & F1 \\
\midrule

\multirow{4}{*}{\textbf{Decoder}} 
& GPT-4.1 Mini & 0.865 & 0.284 & 0.380 & 0.812 & 0.192 & 0.315 & 0.865 & 0.284 & 0.314 \\
& LLaMA-3-8B-Instruct & 0.224 & 0.938 & 0.259 & 0.306 & 0.960 & 0.385 & 0.336 & 0.946 & 0.433 \\
& + CoT & 0.626 & 0.388 & 0.230 & 0.559 & 0.280 & 0.223 & 0.561 & 0.318 & 0.279 \\
& + LoRA & 0.712 & 0.333 & 0.273 & 0.622 & 0.190 & 0.160 & 0.631 & 0.219 & 0.255 \\
\midrule

\multirow{4}{*}{\textbf{Encoder}} 
& RoBERTa + Logistic Regression & 0.750 & 0.214 & 0.194 & 0.840 & 0.285 & 0.330 & 0.510 & 0.364 & 0.329 \\
& RoBERTa + MLP Head & 0.780 & 0.357 & 0.313 & 0.820 & 0.214 & 0.250 & 0.660 & 0.333 & 0.393 \\
& RoBERTa + XGBoost Head & 0.830 & 0.214 & 0.261 & 0.890 & 0.286 & 0.421 & 0.570 & 0.242 & 0.271 \\
& RoBERTa + LoRA + MLP & 0.640 & 0.786 & 0.379 & 0.780 & 0.643 & 0.450 & 0.720 & 0.546 & 0.563 \\

\bottomrule
\end{tabular}
\caption{Overall Performance comparison (\textbf{Accuracy}, \textbf{Recall}, and \textbf{F1}) on three binary classification tasks: Depression, PTSD, and Anxiety.}
\label{tab:main}
\end{table*}

We present the experimental detail and report the corresponding results in this section.
\subsection{5.1 Experimental configuration}
To mitigate output bias and performance degradation caused by label imbalance, we apply upsampling when training decoder-based models, ensuring balanced supervision across classes. When PEFT module is equipped (method 2 and method 3), we experiment with varying LoRA ranks (8, 16, 32) to assess parameter-efficiency and generalization. The dataset is split by user into 80\% for training and 20\% for testing. All models are trained using the AdamW optimizer with a batch size of 8 and a learning rate of $2 \times 10^{-5}$. Experiments are conducted on a Linux server equipped with a single NVIDIA A100 GPU.

\subsection{5.2 Evaluation Metrics}
To assess model performance on mental health diagnosis tasks, we report three core evaluation metrics: \textbf{accuracy}, \textbf{recall}, and \textbf{F1 score}~\cite{tran2017predicting}. \textit{Accuracy} captures the overall proportion of correct predictions and provides a general indication of model reliability across all classes. However, given the clinical relevance of early identification, we include \textit{recall}, the proportion of true positive cases correctly identified, as a primary metric of interest. High recall is critical in mental health contexts, where false negatives may result in delayed or missed intervention.The \textit{F1 score}, defined as the harmonic mean of precision and recall, offers a more balanced view of performance under class imbalance and helps quantify trade-offs between over- and under-diagnosis. 

All metrics are computed individually for each target condition—\textbf{major depressive episodes (MDE)}, \textbf{PTSD}, and \textbf{anxiety disorders}—on a held-out user set, ensuring consistency and fairness across methods. Together, these metrics allow for a clinically meaningful and statistically robust evaluation of diagnostic prediction quality.
We evaluate all three methods, zero-shot prompting with foundation language models, and embedding-based classifier on interview data acrosee three target mental health disorders. Each Each approach is assessed using accuracy, recall, and F1 score, with a particular emphasis on recall due to its clinical relevance in minimizing missed diagnoses.

\subsection{5.3 Overall Evaluation Results}
Table~\ref{tab:main} presents the performance of both decoder-based (Method 1\&2) and encoder-based models (Method 3) across three mental health condition diagnosis: depression, PTSD, and anxiety. Among decoder-based methods, \texttt{GPT-4.1 Mini} achieves the highest accuracy overall, but suffers from low recall, indicating limited sensitivity to positive cases. In contrast, \texttt{LLaMA-3-8B-Instruct} exhibits extremely high recall but poor accuracy and F1, suggesting over-prediction of the positive class under severe label imbalance. The use of Chain-of-Thought (CoT) prompting or LoRA fine-tuning moderately improves the balance between precision and recall.

Encoder-based models, particularly those enhanced with LoRA and MLP heads, demonstrate more stable and balanced performance across all tasks. Notably, \texttt{RoBERTa + LoRA + MLP} achieves the highest F1 scores in PTSD and anxiety detection, indicating its effectiveness in domain-specific adaptation with minimal parameter overhead. Overall, encoder-based approaches with PEFT outperform decoder-based generation methods in terms of classification robustness under label imbalance.

\section{6. Result Analysis and Ablation Studies}
\subsubsection{6.1 Multi-Disorder Inference via Direct Prompting}
We evaluated the zero-shot diagnostic performance of the Meta-LLaMA-3-8B-Instruct model across varying chunk sizes (512, 1024, 2048 tokens) for three mental health conditions: depression, PTSD, and anxiety. The results, summarized in Tables~\ref{tab:llama3_three_disorders} reveal a consistent pattern: the model demonstrates high recall (clinical sensitivity) across all tasks, with values typically above 0.90, but suffers from low F1 scores and overall accuracy, both of which remain below 0.45.

\begin{table}[H]

\centering
\begin{adjustbox}{width=0.45\textwidth}
\begin{tabular}{c|ccc|ccc|ccc}
\toprule
\multirow[c]{2}{*}{\textbf{Model}} 
 & \multicolumn{3}{c|}{\textbf{Depression}} & \multicolumn{3}{c|}{\textbf{PTSD}} & \multicolumn{3}{c}{\textbf{Anxiety}} \\
\cmidrule(lr){2-4} \cmidrule(lr){5-7} \cmidrule(lr){8-10}
& Recall & F1 & Accuracy & Recall & F1 & Accuracy & Recall & F1 & Accuracy \\
\midrule
LLaMA-3 512 tokens  & \textbf{0.950} & 0.247 & 0.163 & \textbf{0.980} & 0.373 & 0.251 & \textbf{0.973} & 0.422 & 0.286 \\
LLaMA-3 1024 tokens & 0.938 & 0.259 & 0.224 & 0.960 & \textbf{0.385} & 0.306 & 0.946 & \textbf{0.433} & 0.336 \\
LLaMA-3 2048 tokens & 0.850 & \textbf{0.260} & \textbf{0.300} & 0.856 & 0.377 & \textbf{0.360} & 0.797 & 0.399 & \textbf{0.358} \\
\bottomrule
\end{tabular}
\end{adjustbox}
\caption{Zero-shot performance of \textbf{Meta-LLaMA-3-8B-Instruct} across chunk sizes for three mental health disorders.}
\label{tab:llama3_three_disorders}
\end{table}

For depression, the model achieved its highest recall of 0.950 using 512-token input, capturing nearly all true positive cases. However, F1 score and accuracy improved modestly with longer context windows, peaking at 0.260 and 0.300, respectively, under the 2048-token settingstill well below optimal thresholds.\\
For PTSD, the recall again peaked at 0.980 with 512-token input. The best F1 score (0.385) was observed at 1024 tokens, while the highest accuracy (0.360) was achieved at 2048 tokens, suggesting a gradual trade-off between sensitivity and overall precision as chunk size increases.\\
For anxiety, the model exhibited slightly more balanced performance. While recall gained the worst score of 0.797 under with the longest chunk size , both F1 and accuracy improved compared to the other disorders. The highest F1 score (0.433) occurred with 1024 tokens, and the best accuracy (0.358) was obtained at 2048 tokens.

\subsection{6.2 LoRA Fine-Tuned Models}
Low-Rank Adaptation (LoRA) has emerged as an efficient fine-tuning strategy for large language models, enabling substantial parameter savings while maintaining strong downstream performance. In our experiments, we tested three LoRA ranks (8, 16, 32) across two architectures: an encoder-only transformer (RoBERTa) and a decoder-only transformer (Meta-LLaMA), evaluating their effectiveness in predicting three mental health conditions.
Overall, RoBERTa consistently outperformed Meta-LLaMA across most metrics. While accuracy generally improved with increasing rank, recall and F1 scores were often higher at lower ranks, though the trend was not strictly monotonic (See Tabel~\ref{tab:lora_finetune}).
\begin{table}[H]

\centering
\scriptsize
\begin{adjustbox}{width=0.48\textwidth}
\begin{tabular}{c|ccc|ccc|ccc}
\toprule
\multirow{2}{*}\textbf{Model} & \multicolumn{3}{c|}{\textbf{Depression}} & \multicolumn{3}{c|}{\textbf{PTSD}} & \multicolumn{3}{c}{\textbf{Anxiety}} \\
\cmidrule(lr){2-4} \cmidrule(lr){5-7} \cmidrule(lr){8-10}
& Acc & Recall & F1 & Acc & Recall & F1 & Acc & Recall & F1 \\
\midrule
\rowcolor{pink!20}
LoRA\_RoBERTa Rank=8  & 0.560 & \textbf{0.857} & 0.353 & 0.700 & \textbf{0.643} & 0.375 & 0.700 & \textbf{0.643} & 0.375 \\
\rowcolor{pink!20}
LoRA\_RoBERTa Rank=16 & \textbf{0.770} & 0.286 & 0.258 & 0.780 & \textbf{0.643} & \textbf{0.450} & \textbf{0.720} & 0.546 & \textbf{0.563} \\
\rowcolor{pink!20}
LoRA\_RoBERTa Rank=32 & 0.640 & 0.786 & \textbf{0.379} & \textbf{0.790} & 0.571 & 0.432 & 0.690 & 0.485 & 0.508 \\
LoRA\_Meta Rank=8     & 0.530 & 0.333 & 0.188 & 0.503 & 0.333 & 0.237 & 0.586 & 0.500 & 0.410 \\
LoRA\_Meta Rank=16    & 0.676 & 0.167 & 0.143 & 0.631 & 0.190 & 0.163 & 0.541 & 0.562 & 0.414 \\
LoRA\_Meta Rank=32    & 0.559 & 0.333 & 0.197 & 0.676 & 0.286 & 0.250 & 0.495 & 0.312 & 0.263 \\
\bottomrule
\end{tabular}
\end{adjustbox}
\caption{LoRA rank impact on encoder and decoder models}
\label{tab:lora_finetune}
\end{table}
\vspace{-0.8em}
\noindent  The performance of depression, the highest accuracy (0.770) was observed with RoBERTa at rank 16. However, both recall and F1 under this configuration remained below 0.3. The highest recall (0.857), also the highest among all tasks—was achieved at rank 8, while F1 peaked at 0.379 with rank 32. \\
In terms of PTSD, accuracy remained above 0.70 across all ranks, with a peak of 0.790 at rank 16. Recall was stronger at lower ranks, whereas the highest F1 score (0.450) again appeared at rank 16, indicating its suitability for this task.\\
For anxiety, both accuracy and recall remained relatively stable across all ranks, with accuracy ranging from 0.541 to 0.720 and recall from 0.500 to 0.6429. The highest accuracy (0.720) was achieved at rank 6, and the highest recall (0.6429) was observed at both ranks 6 and 8. The F1 score peaked at \textbf{0.5625} with rank 6, which was also the highest F1 score across all disorders and configurations. 

\subsection{6.3 Embedding-Based Classifiers}
To complement large-scale language models, we additionally evaluated embedding-based classifiers using sentence embeddings extracted from pretrained \texttt{RoBERTa-base} and \texttt{all-roberta-large-v1} models. As shown in Table~\ref{tab:embedding_classifiers}, across all tasks and models, we observed that \texttt{all-roberta-large-v1} consistently outperformed \texttt{roberta-base} in both recall and F1 score. While overall recall remained low across classifiers—typically below 0.45, accuracy scores were relatively high, frequently exceeding 0.80. 

\begin{table}[H]

\centering
\scriptsize
\begin{adjustbox}{width=0.48\textwidth}
\begin{tabular}{c|ccc|ccc|ccc}
\toprule
\multirow{2}{*}\textbf{Model} & \multicolumn{3}{c|}{\textbf{Depression}} & \multicolumn{3}{c|}{\textbf{PTSD}} & \multicolumn{3}{c}{\textbf{Anxiety}} \\
\cmidrule(lr){2-4} \cmidrule(lr){5-7} \cmidrule(lr){8-10}
& Recall & F1 & Acc & Recall & F1 & Acc & Recall & F1 & Acc \\
\midrule
RoBERTa + LR       & 0.214 & 0.194 & 0.750 & 0.285 & 0.330 & 0.840 & \textbf{0.364} & 0.329 & 0.510 \\
RoBERTa + MLP      & 0.071 & 0.111 & 0.840 & 0.210 & 0.250 & 0.820 & 0.182 & 0.182 & 0.640 \\
RoBERTa + XGBoost  & 0.071 & 0.110 & \textbf{0.840} & 0.143 & 0.190 & 0.830 & 0.200 & 0.200 & 0.600 \\
Large + LR         & \textbf{0.429} & 0.267 & 0.670 & 0.286 & 0.296 & 0.810 & 0.333 & 0.373 & 0.630 \\
\rowcolor{pink!20}
Large + MLP        & 0.357 & \textbf{0.313} & 0.780 & 0.214 & 0.250 & 0.820 & 0.333 & \textbf{0.393} & \textbf{0.660} \\
\rowcolor{pink!20}
Large + XGBoost    & 0.214 & 0.261 & 0.830 & \textbf{0.286} & \textbf{0.421} & \textbf{0.890} & 0.242 & 0.271 & 0.570 \\
\bottomrule
\end{tabular}
\end{adjustbox}
\caption{Performance of embedding-based models with different encoder backbone and classification heads on three mental health disorders. "Large" referes to \textit{all-roberta-large-v1}. Best values per column in \textbf{bold}.}
\label{tab:embedding_classifiers}
\end{table}
We report the performance of embedding-based classifiers on depression classification. The highest recall (0.429) was achieved by logistic regression with \texttt{all-roberta-large-v1} embeddings, while the best F1 score (0.313) was obtained by the MLP classifier using the same embeddings. Accuracy was consistently high across models, ranging from 0.67 to 0.84. In contrast, classifiers based on \texttt{roberta-base} embeddings showed substantially lower recall, highlighting the advantage of using larger, more expressive language models.

A similar trend emerged for PTSD classification. XGBoost paired with \texttt{all-roberta-large-v1} embeddings achieved the highest F1 score (0.421) and accuracy (0.89). However, recall remained modest across classifiers, generally below 0.29, except for logistic regression with \texttt{roberta-base} (0.286). These results suggest that while high accuracy is attainable, embedding models may under-identify true PTSD cases, limiting their sensitivity in clinical settings.

In anxiety classification, the MLP classifier using \texttt{all-roberta-large-v1} embeddings yielded the highest F1 score (0.393) and recall (0.333). Accuracy ranged from 0.51 to 0.66 across models. Compared to depression and PTSD, anxiety prediction exhibited more balanced precision–recall trade-offs, particularly with MLP-based architectures, indicating better stability for this diagnostic category.

\section{7. Discussion and Conclusion}
This study presents a systematic evaluation of three methodological paradigms, including zero-shot prompting with foundation models, LoRA fine-tuning, and embedding-based classifiers for predicting depression, PTSD, and anxiety from real-world interview transcripts. While each approach demonstrates distinct strengths and limitations, several critical themes emerge regarding their practical applicability to mental health screening.

To evaluate the practical utility of GPT-4.1 Mini, we benchmarked its zero-shot performance against both open-source foundation models (e.g., Meta-LLaMA-3) and lightweight embedding-based classifiers. Despite achieving high overall accuracy ($\geq$ 0.80), this align with the exsiting studies \cite{chen2025detectingptsdclinicalinterviews,ben2025assessing}. Howerver, GPT-4.1 exhibited substantially lower recall and F1 scores across all disorders, underscoring the trade-off between general discriminative power and clinical sensitivity.

Across all models, accuracy often remained high—even surpassing 0.85 in several configurations—suggesting models can reliably distinguish between diagnosed and undiagnosed individuals at a population level. However, accuracy alone fails to reflect diagnostic usefulness in clinical settings, where missing true cases (i.e., low recall) can delay care or exacerbate harm. This issue is particularly salient in our results: zero-shot Meta-LLaMA models achieved recall rates above 0.90 for all disorders but suffered from low F1 scores and poor precision, indicating frequent false positives, the lower recall align with the work \cite{ravenda2025evidence}. Conversely, embedding-based classifiers showed high accuracy but considerably lower recall, often below 0.3, underscoring their tendency to under-identify true cases.

From a public health perspective, recall can be interpreted as clinical sensitivity—a measure of how well a model detects individuals who actually need care. Given that depression, anxiety, and PTSD often co-occur, even a single positive flag could increase a user’s awareness of their condition, prompting further clinical consultation. Thus, models with high recall—even at the cost of reduced precision—may serve as effective early-warning tools in digital mental health applications.

LoRA fine-tuning offers a strong trade-off between efficiency and performance. RoBERTa models fine-tuned with LoRA achieved the best overall balance of recall and F1 scores, particularly for anxiety classification (F1 = 0.563). Notably, lower-rank configurations (e.g., Rank=8) sometimes outperformed larger ranks on recall, suggesting that parameter-efficient adaptations may be well-suited for sensitive screening tasks without extensive computational costs. Embedding-based models, while simpler and more interpretable, struggled with recall, indicating limitations in their applicability for nuanced, clinical-like tasks.
\subsection{Limitation and Conclusion}
This study has several limitations that warrant discussion. First, the diagnostic labels are imbalanced, with positive cases comprising only around 20\% of the dataset—and sometimes substantially less. This skew introduces challenges for both training stability and model evaluation, as high accuracy may mask low sensitivity to minority classes. Future work could incorporate reweighting strategies or synthetic oversampling to better calibrate predictions.

Second, although general-purpose large language models have demonstrated strong linguistic capabilities, their understanding of mental health–specific discourse remains limited. These models may lack nuanced knowledge of psychiatric terminology, symptom expression, or the pragmatic context in which mental health conversations occur. For instance, they may misinterpret colloquial expressions of distress or overlook subtle indicators of psychological states. Domain-adapted LLMs trained on relevant corpora—such as therapy transcripts or clinical notes—can provide more reliable grounding for tasks like early detection.

Third, the dataset size is relatively small compared to the scale of LLMs being evaluated. Even with parameter-efficient fine-tuning (e.g., LoRA), limited training samples constrain the model’s ability to generalize, particularly when attempting to update or specialize domain-relevant representations. Freezing core parameters may further amplify this bottleneck.

Fourth, the length of the interviews introduces cognitive strain for decoder-based models with finite context windows. While chunking strategies partially mitigate this issue, they risk discarding context or emphasizing the wrong information. More sophisticated context-aware methods—such as memory-augmented prompting, hierarchical modeling, or relevance-guided chunking—may improve performance.

Looking ahead, several directions hold promise. Beyond general classification, demographic subgroup analyses (e.g., by age, sex, or race) could reveal important disparities in model sensitivity. Incorporating emotional and linguistic signals (e.g., sentiment trajectories, affective intensity) into training data may also enhance predictive validity \cite{gerczuk2023zero,rasool2025emotion}. Finally, analyzing misclassified cases and probing model reasoning pathways through prompt engineering or contrastive examples could reveal failure modes and inform targeted improvements in model alignment.
\section{Ethical Considerations}
While LLM-based personality prediction holds potential for scalable assessment, it raises critical concerns about privacy, consent, and interpretability. We caution against the use of such models in high-stakes decisions without human oversight and advocate for transparent evaluation protocols grounded in psychological theory.


\newpage

\bibliographystyle{aaai}
\bibliography{aaai2026}


\end{document}